\documentclass[runningheads]{llncs}
\usepackage[T1]{fontenc}
\usepackage{graphicx}
\usepackage{times}

\usepackage{multicol}
\usepackage{multirow}
\usepackage[hidelinks]{hyperref}
\usepackage{amsmath}
\usepackage{amssymb}
\usepackage{amsfonts}
\usepackage{textcomp}
\usepackage{xcolor}
\usepackage{epsfig}
\usepackage{eso-pic}
\usepackage{lipsum}
\usepackage{tabularx}
\usepackage{algorithm}
\usepackage{booktabs}
\usepackage{newfloat}
\usepackage{listings}

\usepackage[noend]{algpseudocode}
\usepackage{cite}
\usepackage[english]{babel}

\newcommand{\ourname}{RF-NAV}

\begin{document}
\title{Exploring the Reliability of Foundation Model-Based Frontier Selection in Zero-Shot Object Goal Navigation}
\titlerunning{Exploring the Reliability of Foundation Model-Based Frontier Selection in ZS-OGN}
\author{Shuaihang Yuan\inst{1,2,4} \and
Halil Utku Unlu\inst{3}\and
Hao Huang\inst{2,4}\and
Congcong Wen\inst{2,4}\and
Anthony Tzes\inst{1,2}\and
Yi Fang\inst{1,2,4}\thanks{Yi Fang is the corresponding author with email yfang@nyu.edu}}
\authorrunning{S. Yuan et al.}
\institute{NYUAD Center for Artificial Intelligence and Robotics (CAIR), Abu Dhabi, UAE.
\and New York University Abu Dhabi, Electrical Engineering, Abu Dhabi 129188, UAE.
\and New York University, Electrical \& Computer Engineering Dept., Brooklyn, NY 11201, USA. 
\and Embodied AI and Robotics (AIR) Lab, NYU Abu Dhabi, UAE.\\
}
\maketitle              
\begin{abstract}
In this paper, we present a novel method for reliable frontier selection in Zero-Shot Object Goal Navigation (ZS-OGN), enhancing robotic navigation systems with foundation models to improve commonsense reasoning in indoor environments. Our approach introduces a multi-expert decision framework to address the nonsensical or irrelevant reasoning often seen in foundation model-based systems. The method comprises two key components: Diversified Expert Frontier Analysis (DEFA) and Consensus Decision Making (CDM). DEFA utilizes three expert models—furniture arrangement, room type analysis, and visual scene reasoning—while CDM aggregates their outputs, prioritizing unanimous or majority consensus for more reliable decisions. Demonstrating state-of-the-art performance on the RoboTHOR and HM3D datasets, our method excels at navigating towards untrained objects or goals and outperforms various baselines, showcasing its adaptability to dynamic real-world conditions and superior generalization capabilities.
\end{abstract}

\section{Introduction}
Leveraging foundation models has greatly advanced robotic navigation systems, particularly for frontier-based object goal navigation in indoor environments \cite{yuan2024zero,zheng2208jarvis,zhou2023esc,zhou2023navgpt,yao2022react}. These models enable robots to apply commonsense reasoning during exploration and object search \cite{komorowski2021minkloc3d, xia2021soe, xia2023casspr, yuan2020ross, huang2024noisy}. For instance, when the target is a desk, the robot understands that desks are often paired with chairs. This enhanced perception and reasoning allow navigation systems to more effectively identify promising frontiers for exploration, resulting in higher success rates compared to traditional methods like distance-based and Gaussian-process-based frontier selection \cite{ali2023gp,suzuki2020multi,jadidi2014exploration,aliautonomous,ramakrishnan2022poni,zhou2021fuel}.

Recently, researchers have integrated chain-of-thought (COT) prompting \cite{zhang2022automatic,wei2022chain,yao2023tree} into foundation models to enhance commonsense reasoning in navigation systems \cite{yang2022chain,shah2023navigation,koubaa2023rosgpt,zhou2023esc}. COT generates short, human-like reasoning steps during navigation tasks. For example, when searching for a desk, the system might reason, "A desk is often found in a study room, which typically contains books, laptops, and chairs. If I encounter these objects near an unexplored frontier, I should explore it first." COT enhances navigation performance by offering a more transparent and interpretable decision-making process. However, it often relies on greedy decoding, which can lead to suboptimal reasoning \cite{wei2022chain, wang2022self,chowdhery2023palm}, resulting in nonsensical or irrelevant conclusions and decreasing system reliability \cite{wang2023assessing}. This limitation highlights the need for more advanced methods to enhance the robustness of foundation model-driven navigation systems.

\begin{figure}[t]
    \centering
    \includegraphics[width=0.8\linewidth]{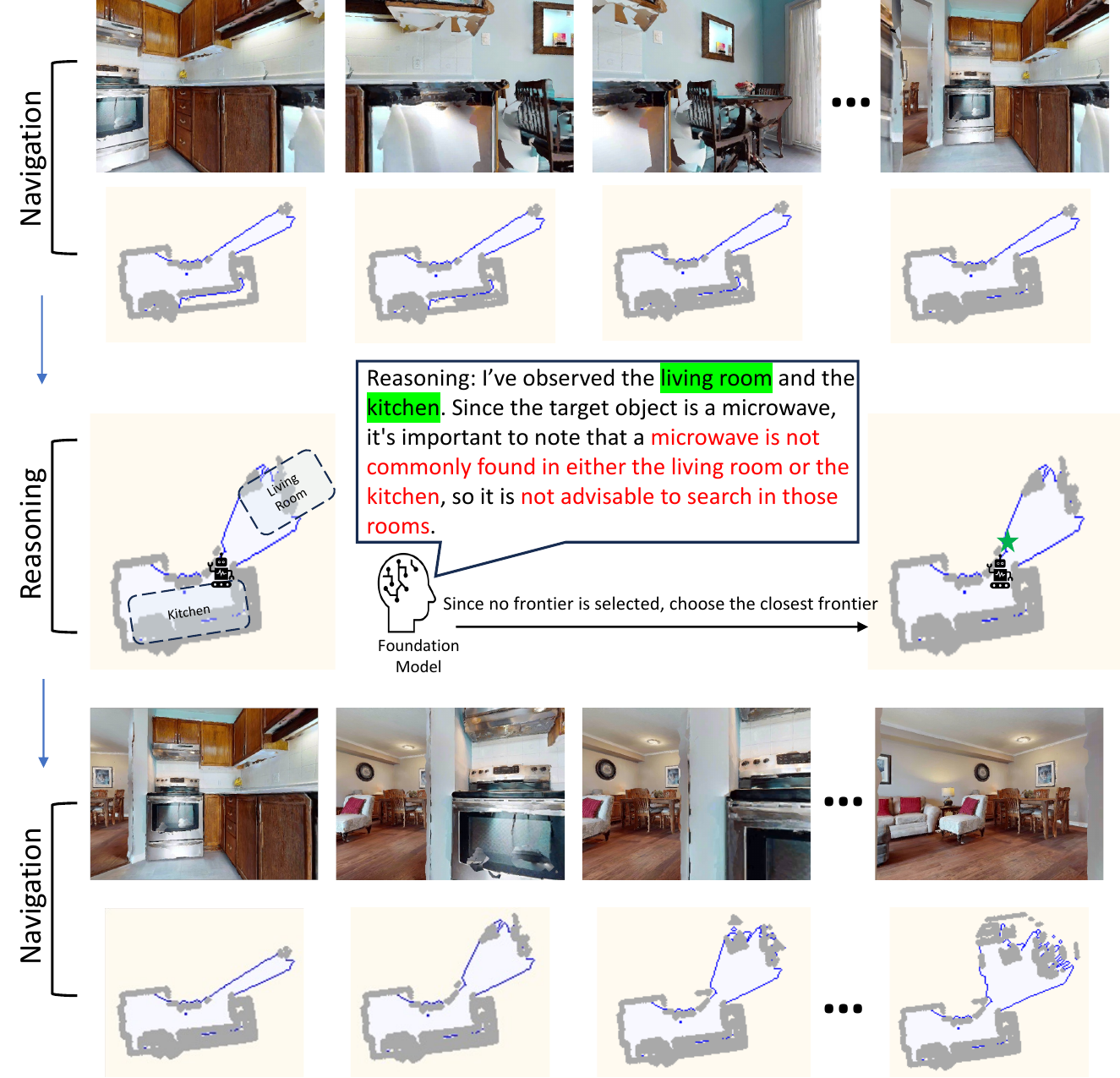}
    \caption{Instances of nonsensical or irrelevant reasoning, during the frontier selection in Zero-Shot Object Goal Navigation. The green text indicates a correct understanding of the scene, while the red text refers to the reasoning that contradicts human intuition. }
    \label{fig:intro}
\end{figure}

In real-world dynamic environments, the reliability of foundation model-driven navigation systems is vital, especially in adapting to changing scenarios where consistent performance is difficult to maintain. Our research addresses this challenge by developing robust commonsense reasoning for zero-shot object goal navigation. This approach is critical for navigating unpredictable conditions where robots encounter previously unseen objects or situations. Unlike traditional tasks, zero-shot navigation \cite{zhou2023esc,gadre2023cows} requires the system to orient toward goals without prior explicit training, demanding advanced generalization and commonsense knowledge to reliably adapt across unfamiliar objects and situations in dynamic environments.

In this paper, we introduce a novel frontier selection method for zero-shot object goal navigation (ZS-OGN). The method features two key components, with the first being Diversified Expert Frontier Analysis (DEFA), inspired by Portfolio Theory \cite{constantinides1995portfolio,markowitz1991foundations,markowitz2010portfolio,mangram2013simplified}. DEFA leverages the expertise of three foundation models, each serving as an expert in a specific aspect of frontier selection. The first expert focuses on selecting frontiers based on furniture arrangements, such as identifying desk-like setups with chairs. The second expert prioritizes frontiers leading to rooms where the target object, like a desk, is more likely to be found, such as study rooms. The third expert uses visual observation to apply commonsense reasoning dynamically based on the scene.

We introduce Consensus Decision Making (CDM) as the second component for frontier selection, inspired by self-consistency. CDM first seeks unanimous expert approval, and if not achieved, selects a frontier endorsed by at least two experts. This approach balances the diversity of DEFA experts while enhancing reliability in zero-shot object goal navigation. Our system's state-of-the-art performance on the RoboTHOR \cite{deitke2020robothor} and HM3D \cite{ramakrishnan2021hm3d} datasets validates its effectiveness, with detailed analysis further demonstrating the reliability of our method compared to baseline approaches.

\section{Related Work}
\subsection{Language-driven Zero-shot Object Navigation} 
In Object Goal Navigation (OGN), the goal is to efficiently explore a new environment while searching for a non-visible target object. Previous research often relies on visual context through imitation~\cite{silver2008high,karnan2022voila} or reinforcement learning~\cite{kahn2018self,wohlke2021hierarchies}, which require extensive data collection and annotations, limiting their practicality in real-world environments. The focus has shifted towards zero-shot object navigation, enabling robots to adapt to new objects and environments without specific training~\cite{zhao2023zero,majumdar2022zson,dorbala2023can,zhao2023semantic}. Clip-Nav~\cite{dorbala2022clip} and CoW~\cite{gadre2022clip} use CLIP~\cite{radford2021learning} for zero-shot navigation, while L-ZSON~\cite{gadre2023cows} employs Frontier-Based Exploration (FBE)~\cite{yamauchi1997frontier} to navigate between known and unknown spaces, outperforming learning-based methods~\cite{schulman2017proximal,wijmans2019dd}. Unlike recent works~\cite{wu2024voronav,cai2023bridging} that train policy networks for frontier exploration, we leverage Large Language Models (LLMs) like GPT-3.5~\cite{openai2023chatgpt} and GPT-4~\cite{openai2023gpt4} to make navigational decisions directly, bypassing the need for any training process.

\subsection{Commonsense Reasoning in Navigation}

Commonsense reasoning~\cite{kojima2022large,krause2023commonsense} is critical for achieving human-like intelligence in robotics~\cite{song2023llm,huang2022language}. Large pre-trained LLMs with reasoning capabilities are becoming increasingly vital for navigation. For instance, BERT~\cite{devlin2018bert} enhances navigation by linking language instructions to navigational paths \cite{majumdar2020improving}, while GPT-4 further improves commonsense reasoning in navigation~\cite{dorbala2023can,zhou2023navgpt,chen2023not,shah2023lm}. NavGPT \cite{zhou2023navgpt} integrates prompt-based methods, like ReAct \cite{yao2022react}, with discrete action spaces for better navigation. Other work leverages commonsense knowledge and semantic mapping to improve goal identification and navigation~\cite{chaplot2020learning,chaplot2020object,chen2022weakly}. Recent research also integrates predictions from language models with planning or probabilistic inference~\cite{shah2023lm,huang2023grounded}, while some focus on grounding language models in image observations~\cite{jiang2022vima,huang2023language,driess2023palm}. JARVIS~\cite{zheng2208jarvis} offers a neuro-symbolic framework for generalizable conversational embodied agents, while ESC~\cite{zhou2023esc} pre-computes object-room relationships for zero-shot navigation, though it struggles with evolving environments. We propose a novel approach that dynamically infers commonsense knowledge from observed scenes, overcoming this limitation.

\begin{figure*}[t]
    \centering
    \includegraphics[width=0.9\textwidth]{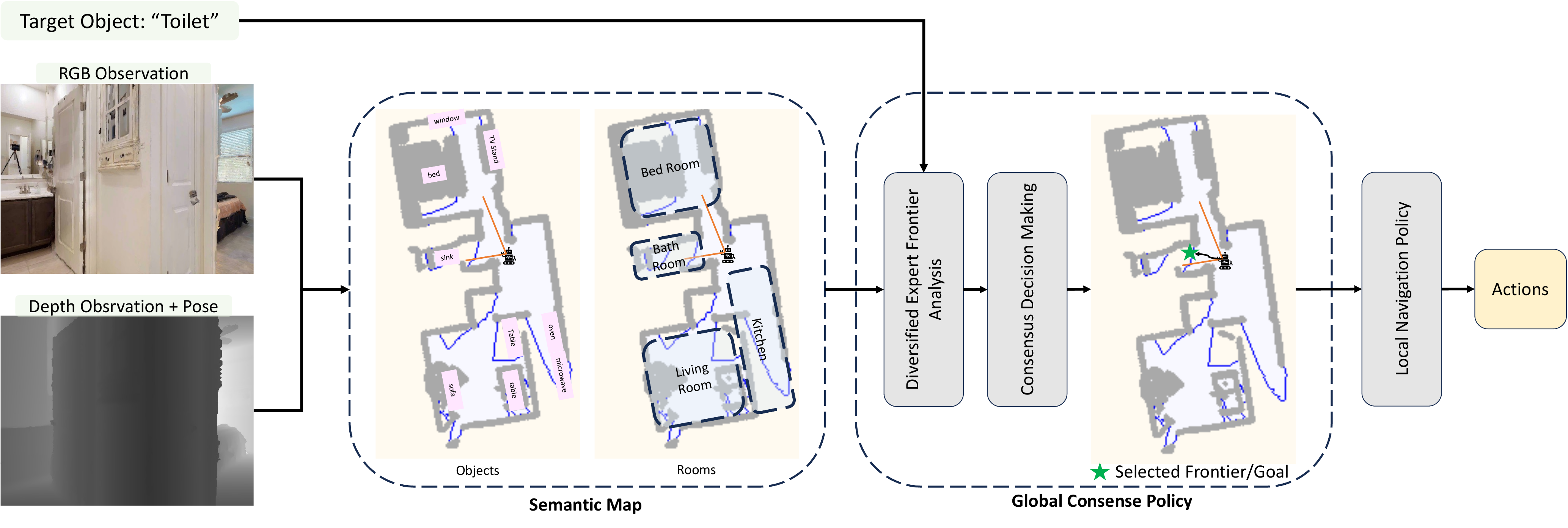}
    \caption{Workflow of the proposed ZS-OGN system, \ourname, for Zero-Shot Object Goal Navigation (ZS-OGN). The process begins with RGB and depth observations leading to the creation of a semantic map, which includes identified objects and room labels. This map informs the Diversified Expert Frontier Analysis (DEFA) and subsequent Consensus Decision Making (CDM) to select the most viable frontier or goal, here exemplified by the search for a 'Toilet.' The chosen goal is then fed into the Local Navigation Policy, which determines the actions necessary for the robot to explore the unknown environment.}
    \label{fig:pipe}
\end{figure*}

\section{Problem Formulation}

In ZS-OGN, the robot must navigate to a target object \( g_i \) in an unfamiliar environment \( s_i \), without prior training on navigation data. Each episode is defined as \(\mathcal{E}_i=\{g_i,s_i,p_0\}\), where \( p_0 \) is the robot's starting position. At each step \( t \), the robot receives an observation \(\mathcal{O}_t=\{I_t,d_t,x_t,y_t,\theta_t\}\), which includes a color image \( I_t \), depth image \( d_t \), and its pose (position \( (x_t, y_t) \) and orientation \(\theta_t\)). Over time, the robot accumulates pose readings to track its relative position \( p_t \). Based on these observations, the robot selects an action \( a \) from the action space \(\mathcal{A}=\{\text{``move forward", ``turn left", ``turn right", ``stop"}\}\) via a policy function \(\pi(\cdot)\). Success is achieved if the robot executes the "stop" action within a predefined distance of the target object. In this study, we frame the navigation task as a sequence of decisions starting at time step \( t=0 \) and ending at the final step \( T \), either when the target is found or the maximum steps are reached. The challenge lies in developing a zero-shot policy \(\pi\), designed to select the optimal action \( a_t \) at each step \( t \) based on the observation \(\mathcal{O}_t\).

\section{Method}
To tackle the problem outlined above, we employ Frontier-based Exploration \cite{yamauchi1997frontier}. Our method is organized into three modules: Mapping, Global Commonsense Policy, and Local Navigation Policy, as illustrated in Figure~\ref{fig:pipe}. First, \ourname constructs a semantic and frontier map based on the observation \(\mathcal{O}_t\) (see Sec. \ref{sec:map}). Next, in the Global Commonsense Policy, we introduce Diversified Expert Frontier Analysis (DEFA) and Consensus Decision Making (CDM) to select the most promising frontier for further exploration (see Sec. \ref{sec:global_policy}). Finally, the Local Navigation Policy plans the path to the frontier or target object and generates the necessary actions to reach it, as detailed in Section~\ref{sec:local_policy}.

\subsection{Mapping}
\label{sec:map}
Constructing semantic and frontier maps are fundamental modules in various frontier-based navigation systems \cite{ramakrishnan2022poni}. Following the approach outlined in \cite{chaplot2020object}, we construct the semantic map using RGB-D images and the agent's pose. The RGB-D input is transformed into 3D voxels 
and then projected onto a top-down 2D navigation map. We utilize the ESC \cite{zhou2023esc} framework to extract semantic information, including common objects and room types in \(\mathcal{E}_i\), using the Grounded Language-Image Pre-training (GLIP) model \cite{li2022grounded}. This model enables zero-shot detection capabilities through natural language prompting, allowing us to detect both objects and room types. The zero-shot detection process using object prompting \(P_{o}\) and room prompting \(P_r\) is formally described as:
\( \{o_{t,n}, b_{o_{t,n}}\} = GLIP(I_t, P_o), \{r_{t,n}, b_{r_{t,n}}\} = GLIP(I_t, P_r) \). Here, \(o_{t,i}\) and \(r_{t,i}\) represent the predicted labels of objects and rooms, respectively, while \(b_{o_{t,i}}\) and \(b_{r_{t,i}}\) denote their bounding boxes. The index \(i\) indicates the detected room or object at step \(t\). The locations of detected rooms and objects are then projected to form the semantic map.

\begin{algorithm*}
\caption{Consensus Decision Making}
\label{alg:cdm}
\begin{algorithmic}[1]
\State \textbf{Input:} Sets $F_{\text{O2F}}$, $F_{\text{R2F}}$, $F_{\text{SLE}}$ of frontiers recommended by three experts; frontier distance matrix $d$ \Comment{Define inputs}
\State \textbf{Output:} A frontier \Comment{Define output}

\Procedure{FindConsensus}{$F_{\text{O2F}}, F_{\text{R2F}}, F_{\text{SLE}}$} \Comment{Begin consensus finding}
    \State $F_{\text{unanimous}} \gets F_{\text{O2F}} \cap F_{\text{R2F}} \cap F_{\text{SLE}}$ \Comment{Check for unanimous consensus}
    \If{$F_{\text{unanimous}} \neq \emptyset$}
        \State \textbf{return} $F_{\text{unanimous}}$ \Comment{Return if unanimous consensus exists}
    \EndIf
    \State $F_{\text{partial}} \gets (F_{\text{O2F}} \cap F_{\text{R2F}}) \cup (F_{\text{O2F}} \cap F_{\text{SLE}}) \cup (F_{\text{R2F}} \cap F_{\text{SLE}})$ \Comment{Check for partial consensus}
    \If{$F_{\text{partial}} \neq \emptyset$}
        \State \textbf{return} $F_{\text{partial}}$ \Comment{Return if partial consensus exists}
    \Else
        \State \textbf{return} \Call{ClosestFrontier}{$d$} \Comment{Return the closest frontier}
    \EndIf
\EndProcedure

\Procedure{SelectFrontier}{} \Comment{Begin frontier selection}
    \State $F_{\text{consensus}} \gets$ \Call{FindConsensus}{$F_{\text{object}}, F_{\text{room}}, F_{\text{proximity}}$} \Comment{Get consensus set}
    \State $F_{\text{selected}} \gets \arg\max_{f \in F_{\text{consensus}}} d[f]$ \Comment{Select max confidence frontier}
    \State \textbf{return} $f_{\text{selected}}$ \Comment{Return selected frontier}
\EndProcedure

\State $f_{\text{final}} \gets$ \Call{SelectFrontier}{} \Comment{Determine final selection}
\State \textbf{Output} $f_{\text{final}}$ \Comment{Output the final selected frontier}

\end{algorithmic}
\end{algorithm*}

To generate the frontier map from the navigation map, we adhere to the methodology outlined in \cite{ramakrishnan2022poni}. Initially, we identify the edges of the free area, defined as the space visible to the agent and not obstructed by obstacles. Subsequently, the boundary points between the free area and the unexplored space are identified as candidate frontiers. These candidates are then sent to the Global Commonsense Policy to determine the most promising frontier for the next phase of exploration.

\subsection{Global Commonsense Policy}
\label{sec:global_policy}
Global Commonsense Policy, \(\pi_{\text{global}}\), is responsible for selecting
the best frontier
by leveraging the commonsense reasoning ability from the foundation models. The selection of a frontier is based on the nearby objects, the room type, and the room configuration. The output is a chosen frontier \(f_t\), which is a point in the environment the robot aims to reach. Global Commonsense Policy
consists of two components: (1) Diversified Expert Frontier Analysis (DEFA) to analyze the frontiers from diverse perspectives and (2) Consensus Decision Making (CDM) to produce the final decision by considering all the options produced by the DEFA.

\subsubsection{Diversified Expert Frontier Analysis} In the DEFA module, we employ three distinct expert models to realize the decision-making process in ZS-OGN, each bringing a unique perspective to frontier selection. The Object2Frontier Expert (O2F) specializes in analyzing the objects near potential frontiers, identifying frontiers that are indicative of the target object's likely presence. We leverage the reasoning ability from an LLM to realize this, as \( F_{O2F}(\{o\}) \rightarrow \{S_{O2F}\} \), where \( \{o\} \) is the set of observed objects near each frontier, and \( S_{O2F} \) is the selected frontiers. We use ChatGPT-3.5 as the O2F in all of our experiments.

In addition to the O2F, we further employ an LLM as the Room2Frontier Expert (R2F). R2F assesses the room type associated with each frontier, prioritizing those that align with the expected location of the target, such as study rooms for a desk, denoted as \( F_{R2F}(\{r\}) \rightarrow S_{R2F} \), with \( r \) denoting room types and \( S_{R2F} \) the selected frontier by this expert. We also adopt the ChatGPT-3.5 as the R2F throughout our experiments.

Lastly, to complement the analysis, we adopt a Scene Layout Expert (SLE) specifically to compensate for the loss of visual information not addressed by the previous experts. This expert leverages visual data to dynamically reason commonsense knowledge based on the observed scene. The SLE is implemented using a Multimodal Large Language Model (MLLM), GPT-4V, which processes RGB observations \(\{I\}\), alongside detected objects and room types. This is formulated as \( F_{SLE}(\{I\}, \{o\}, \{r\}) \rightarrow S_{SLE} \). Each expert operates independently, diversifying the frontier evaluation criteria and determining a reliable frontier to navigate. To integrate the decisions from various experts, we utilize the Consensus Decision Making component.

\subsubsection{Consensus Decision Making}
We introduce Consensus Decision Making (CDM) for selecting the frontier from all recommendations from the experts. This straightforward yet effective approach relies on majority voting and reduces the occurrence of instances of nonsensical or irrelevant reasoning. Ideally, all experts agree upon a single frontier. 
When a unanimous selection is not achieved, the strategy chooses the frontier endorsed by the majority.
After the selection is made, we rank the frontiers according to their distances to the robot's current location to determine the final selection.

While we have demonstrated that the lower bound of our method to produce an irrational result is lower than that of relying on a single expert to determine the frontier, it is important to note that our algorithm might still encounter situations where the three experts do not reach any consensus. To mitigate this issue, we have incorporated a fallback strategy, in which the robot will select the closest frontier if no  consensus is achieved. We detail this entire Consensus Decision-Making (CDM) process in Algorithm~\ref{alg:cdm}. After the goal (either the frontier or the location of the target object) has been selected, the Local Navigation Policy generates the path planning and sequences the actions needed to reach the goal.

\begin{algorithm}
\caption{Frontier-based Exploration Method}
\label{alg:pipe}
\label{alg:frontier_exploration}
\begin{algorithmic}[1]

\Require Observation $\mathcal{O}_t$, Semantic Map $\mathcal{M}_{\text{semantic}}$, Frontier Map $\mathcal{M}_{\text{frontier}}$

\State \textbf{Mapping:}
\State Detect objects and room types using GLIP model
\State Construct a semantic map using RGB-D images and the agent's position
\State Construct a frontier map and identify candidate frontiers
\State \textbf{End Mapping}

\State \textbf{Global Commonsense Policy} $\pi_{\text{global}}$:
\State \textit{Diversified Expert Frontier Analysis (DEFA):}
\State $S_{O2F} \leftarrow F_{O2F}(\{o\})$ \Comment{Object2Frontier Expert}
\State $S_{R2F} \leftarrow F_{R2F}(\{r\})$ \Comment{Room2Frontier Expert}
\State $S_{SLE} \leftarrow F_{SLE}(\{I\}, \{o\}, \{r\})$ \Comment{Scene Layout Expert}

\State \textit{Consensus Decision Making (CDM):}
\State $f_{final} \leftarrow \text{CDM}(S_{O2F}, S_{R2F}, S_{SLE})$
\State \textbf{End Global Commonsense Policy}

\State \textbf{Local Navigation Policy} $\pi_{\text{local}}$:
\State Plan path to $f_{final}$ using Fast Marching Method (FMM)
\State $a_t \leftarrow \pi_{\text{local}}(\mathcal{O}_t, f_{final})$
\State \textbf{End Local Navigation Policy}

\end{algorithmic}
\end{algorithm}

\subsection{Local Navigation Policy}
\label{sec:local_policy}
To navigate from the agent’s current location to a goal produced from the CDM, we employ the Fast Marching Method (FMM) \cite{sethian1996fast}, a numerical technique that efficiently solves the Eikonal equation, providing a way to estimate the minimal time necessary for the agent to reach the selected frontier from its starting point in the environment. Once a frontier is selected, the local navigation policy $\pi_{\text{local}}$ is responsible for planning the path to this frontier and generating the appropriate actions to navigate along this path. This network takes as input the current observation \(\mathcal{O}_t\) and the selected frontier \(f_{final}\), and outputs the action \(a_t\) to be taken at time step \(t\).
   \(a_t = \pi_{\text{local}}(\mathcal{O}_t, f_{final}) \). 
The combined policy \(\pi\) operates by first using \(\pi_{\text{global}}\) to select a frontier and then using \(\pi_{\text{local}}\) to navigate towards this frontier. This process is repeated at each time step \(t\) until the robot either reaches the target object or the episode ends.

In this formulation, \(\pi_{\text{global}}\) provides a strategic decision-making capability, selecting waypoints or goals that guide the overall navigation task. In contrast, \(\pi_{\text{local}}\) is focused on the immediate, tactical decisions required to navigate safely and efficiently to the chosen frontier. This division allows the policy to effectively manage both the high-level navigation objectives and the detailed, moment-to-moment challenges of robot movement in an unknown environment. The formulation of our complete navigation system flow can be found in Algorithm~\ref{alg:pipe}

\section{Simulation Studies}

\subsection{Datasets and Metrics}
\textbf{HM3D} \cite{ramakrishnan2021hm3d}, a foundational dataset for the Habitat 2022 ObjectNav challenge, includes 142,646 object instances across 40 classes and 216 3D environments, covering 3,100 rooms. We follow prior validation settings \cite{zhou2023esc,gadre2023cows} to evaluate our method. \textbf{RoboTHOR} \cite{deitke2020robothor} serves as a real-world benchmark with 89 apartment scenes and 731 unique objects. We assess our method on 1,800 validation episodes across 15 environments, focusing on 12 target object categories for zero-shot object goal navigation. 

We use Success Rate (SR) and Success Weighted by Path Length (SPL) to evaluate the effectiveness of our proposed method. \textbf{SR} metric focuses on the agent's accuracy in reaching the designated target, expressed as a percentage, where a higher value indicates better performance. SR is a binary indicator of whether the robot successfully stops within 0.1m of the target object \(g_i\) within the episode. In addition, we also measure SPL.

\textbf{SPL} is a metric that evaluates success relative to the shortest possible path, normalized by the actual path taken by the agent. It effectively measures the efficiency of the agent's success in reaching its goal.

\subsection{Baselines}
We compare our method with two state-of-the-art (SOTA) approaches in zero-shot object goal navigation (ZS-OGN) and our own baseline methodologies. CoW (CLIP on Wheels) \cite{gadre2023cows} tackles language-driven ZS-OGN without fine-tuning, using CLIP to identify target objects and select frontiers. We also evaluate CoW variants: \textit{CLIP-Ref, CLIP-Patch, CLIP-Grad, MDETR}. ESC~\cite{zhou2023esc} applies commonsense knowledge from a pre-trained LLM to navigate unseen environments, combining vision and language models for object identification and reasoning. Additionally, we developed a baseline where a single expert repeatedly determines the next frontier, selecting the most frequent outcome for exploration, unlike the more sophisticated reasoning process in ESC.

\subsection{Results on HM3D}

In this dataset, our ZS-OGN system outperforms the CoW and ESC models in both SPL and SR metrics, as shown in Table \ref{tab:model_comparison}. The SR improvement from 35.4 to 37.4 highlights our model's enhanced understanding of environmental semantics, aided by the Multimodal Large Language Model expert. The SPL increase from 17.8 to 21.7 demonstrates the effectiveness of our multi-expert approach in exploring unknown environments. Additionally, our model surpasses the Ours (M.V) approach, which relies on a single expert's majority consensus. The collaborative decision-making process in our model results in a more refined navigation strategy, leading to higher SPL in complex real-world HM3D environments.

\subsection{Results on RoboTHOR}
In this dataset, we test our ZS-OGN system in an unknown environment, where it outperforms the CoW and ESC models in both SPL and SR metrics, as shown in Table \ref{tab:model_comparison}. The increase in SR from 35.4 to 37.4 suggests that our model's understanding of environmental semantics, enhanced by the Multimodal Large Language Model expert, significantly improves navigation capabilities. Similarly, the SPL improvement from 17.8 to 21.7 indicates that our model's multi-expert approach is particularly effective in exploring unknown environments. Moreover, the superiority of our proposed frontier selection method is further confirmed by our model's improved performance over
the single expert's majority consensus baseline, indicating that the improvements in our navigation strategy generalize to RoboTHOR's environments as well. 

\setlength{\tabcolsep}{3pt}

\begin{table}[htbp]
\centering
\caption{Comparison of Zero-shot OGN methods on the HM3D and RoboTHOR benchmarks using SPL and SR metrics, showing our models' superior performance, especially with the Consensus Commonsense strategy.}
\begin{tabular}{lccccc}
\toprule
\textbf{Model} & \textbf{Frontier Selection}& \multicolumn{2}{c}{\textbf{HM3D}} & \multicolumn{2}{c}{\textbf{RoboTHOR}} \\
 &  &   \textbf{SPL↑} & \textbf{SR↑} & \textbf{SPL↑} & \textbf{SR↑} \\
\midrule
CLIP-Ref & Closest & - & - & 2.1 &  2.7  \\
MDETR & Closest & - & - & 8.4 &  9.9  \\
CLIP-Grad & Closest & - & - & 9.7 &  13.8  \\
CLIP-Patch & Closest & - & - & 10.6 &  20.3  \\
CoW & Closest & - & - & 16.9 &  26.7  \\
ESC  & Commonsense &  17.8  & 35.4 & 18.2  & 34.5 \\
Ours(k=3)  & Majority Commonsense& 18.9 & 36.3 & 20.6 & 35.2 \\
Ours(k=5)  & Majority Commonsense& 19.1 & 36.6 & 20.8 & 35.6 \\
Ours  & Consensus Commonsense & \textbf{21.7} & \textbf{37.4} & \textbf{22.3} & \textbf{36.8} \\
\bottomrule
\end{tabular}
\label{tab:model_comparison}
\end{table}

\begin{figure}[ht]
  \centering
  \includegraphics[width=0.9\linewidth]{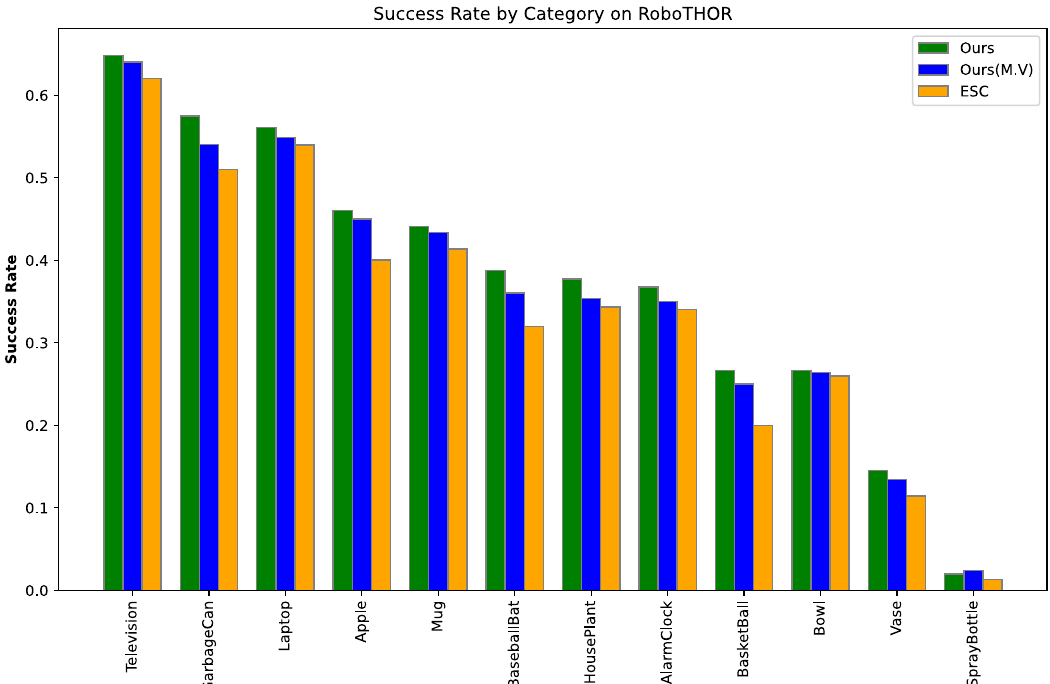}
  \caption{Success rates for ZS-OGN in twelve target goal categories. The comparison is among our proposed method, our baseline method, and ESC \cite{zhou2023esc}
  }
  \label{fig:cat}
\end{figure}
\begin{figure*}[ht]
  \centering
  \includegraphics[width=0.9\textwidth]{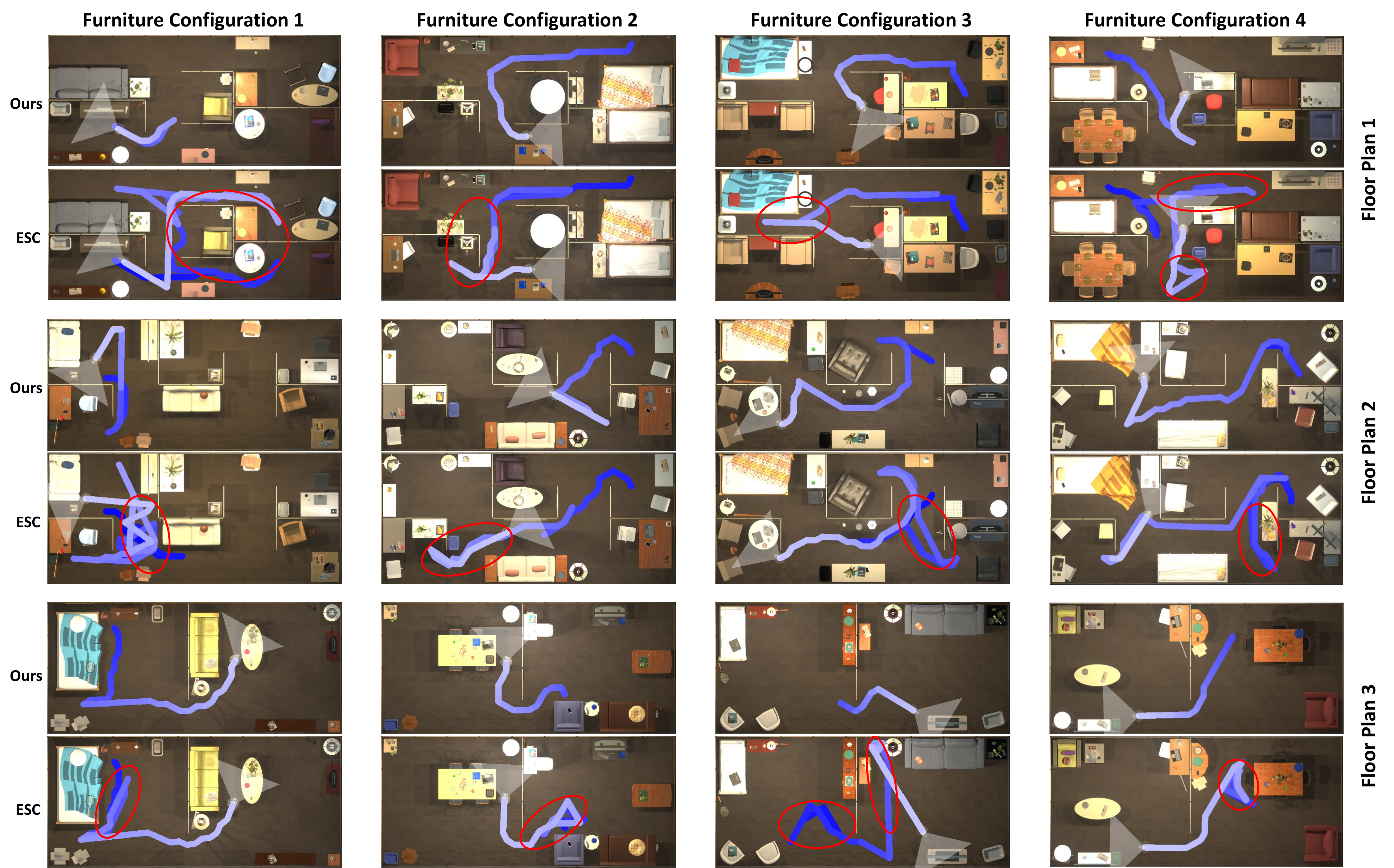}
  \caption{A comparison of the generated paths to target objects between our proposed method and ESC \cite{zhou2023esc}. Paths generated by our proposed method are more direct and efficient. Instances of zigzagging motion are marked in red ellipses.
  }
  \label{fig:path}
\end{figure*}

\begin{figure*}[ht]
  \centering
  \includegraphics[width=0.9\textwidth]{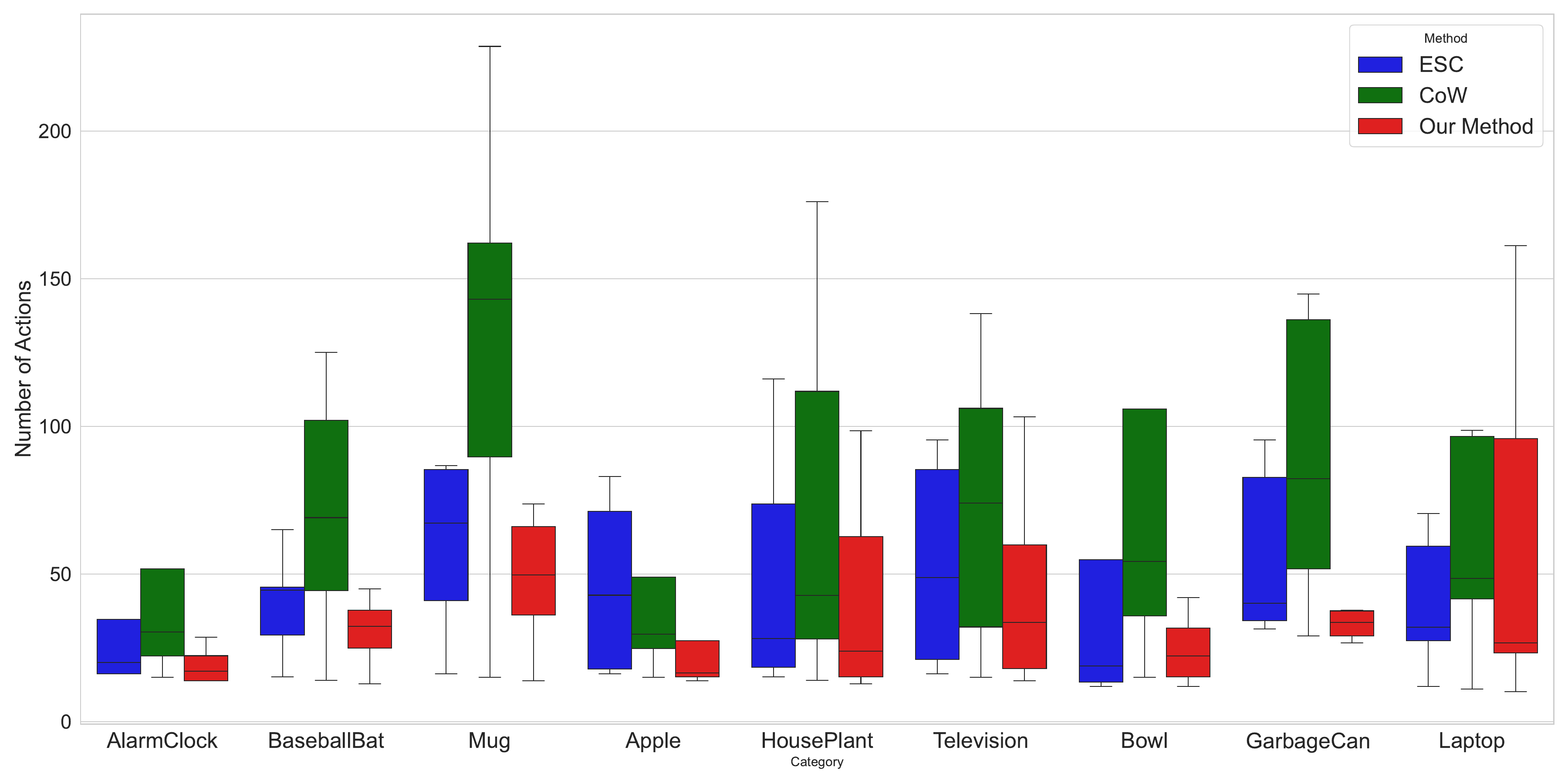}
  \caption{The comparison of the distribution of the average number of actions to complete the zero-shot OGN across different target objects between our method and ESC. \cite{zhou2023esc}
  }
  \label{fig:time}
\end{figure*}

\section{Analysis}
\subsection{Effect of Reliable Frontier Selection}
In this section, we analyze the effectiveness of our proposed frontier selection method within the RoboTHOR environment. For a fair comparison, we evaluate our method alongside ESC and GoW (GLIP on Wheel) \cite{zhou2023esc,gadre2023cows}. GoW is a variant of CoW, utilizing GLIP instead of CLIP. The key difference among these models lies in their frontier selection mechanisms. Our method introduces a novel multi-expert frontier reasoning process combined with an innovative, condensed decision-making approach. In contrast, ESC employs a single expert (GPT-3) for frontier reasoning, while GoW uses a closest frontier strategy.

We visualize the navigation paths of ESC and our method across unknown environments with three room layouts and four object placements each. Figure \ref{fig:path} shows our approach is more reliable and efficient than ESC. In Floor Plan 1, ESC's zigzagging, highlighted by red ellipses, indicates indecision and poor frontier selection, while our method takes a more direct path, demonstrating better reasoning and efficiency. These results emphasize a more effective frontier selection and a navigation strategy that enhances reliability, reduces detours, and shortens goal-reaching time.

We evaluate the number of actions required by robots to locate target objects across categories in unknown environments using RoboTHOR, comparing our method with ESC and GoW. The box plot in Figure \ref{fig:time} shows the median actions, 25th and 75th percentiles, and overall range. Our method consistently achieved lower median values across most categories compared to ESC, indicating higher efficiency, despite greater variability. ESC showed more predictable performance but required more actions, while CoW had fewer outliers due to its nearest-frontier strategy but resulted in higher median values. Overall, our method demonstrated more efficient navigation with fewer actions on average.

\begin{table}[ht]
\centering
\caption{Comparison of models across different metrics.}
\begin{tabular}{lccc}
\toprule
\textbf{Metric} & \textbf{GoW} & \textbf{ESC} & \textbf{Ours} \\
\midrule
FrontierDist (m) & 8.2 & 7.6 & \textbf{6.8} \\
Exploration (\%) & 14.3 & 10.6 & \textbf{9.2} \\
Detection (\%) & 40.6 & 40.8 & \textbf{40.6} \\
Planning (\%) & 12.1 & \textbf{9.5} & 9.6 \\
\bottomrule
\end{tabular}
\label{tab:error_comparison}
\end{table}

\subsection{Error Analysis}
In this section, we conduct an error analysis of our proposed method, ESC and GoW. We follow the standard protocols in \cite{gadre2023cows, zhou2023esc} to analyze three types of error: 
1) \textit{Detection error} happens when the agent either misses the goal or incorrectly believes it has detected the goal. 
2) \textit{Planning error} arises when the agent either recognizes the target but cannot reach it or gets stuck without spotting the goal, reflecting the path-planning ability of the system.
3) \textit{Exploration error} occurs when the agent fails to see the goal object due to issues other than planning or detection, assessing its ability to approach the goal. 

In Table~\ref{tab:error_comparison}, we note that the detection errors for our method, ESC, and GoW are nearly identical, which is expected given that all three methods employ the same detection head. The similarity of detection errors across all methods suggests that enhancing zero-shot object detection models could be a valuable direction for future ZS-OGN research. Regarding planning errors, our method and ESC exhibit similar rates, as both use FMM for path planning, whereas GoW, which employs A*, shows a higher error rate in this specific dataset. Concerning the Exploration Error, our method outperforms ESC, indicating that it more effectively aids the agent in exploring the environment and approaching the object.

\subsection{Effect of Different Experts}
We conducted an experiment to validate the effectiveness of various experts. The results are presented in Table~\ref{tab:expert}, where, for the multi-expert method, we adjusted the consensus decision-making process by only proceeding once both experts concurred.

\begin{table}[ht]
\centering
\caption{Comparison of models across different metrics.}
\begin{tabular}{lcccccc}
\toprule
\textbf{Metric} & \textbf{O2F} & \textbf{R2F} & \textbf{SLE} & \textbf{SLE+R2F}& \textbf{SLE+O2F}& \textbf{O2F+R2F} \\
\midrule
SPL & 17.8 & 18.2 & 19.6 & \textbf{21.7} & \textbf{21.7} & 20.9\\
\bottomrule
\end{tabular}
\label{tab:expert}
\end{table}
The results indicate that visual cues improve the outcome significantly. Notably, SLE+R2F and SLE+O2F exhibit similar performances, which is expected since visual information can provide insights into both room type and object co-occurrence.

\section{Conclusion and Future Work}

Our study introduces an innovative approach to enhance the reliability of foundation model-driven frontier selection for navigation systems, particularly in zero-shot object goal navigation scenarios. By integrating the Diversified Expert Frontier Analysis (DEFA) and Consensus Decision Making (CDM), our method improves commonsense reasoning for frontier selection by diversifying the reasoning and decision-making process. The CDM component, inspired by the concept of self-consistency, further ensures reliability by requiring majority expert agreement for frontier selection. The promising performance on the RoboTHOR and HM3D datasets, along with a comprehensive analysis against various baselines, demonstrate its effectiveness and reliability in zero-shot navigation tasks. While our approach shows great promise, there are opportunities for future improvement. The system's complexity introduces computational demands that can be optimized to enhance real-time performance. Additionally, although decision-making consistency has improved, occasional instances of nonsensical or irrelevant reasoning highlight areas where further refinement can increase reasoning accuracy. These enhancements will be key to advancing the system’s efficiency and reliability, particularly in complex and dynamic environments.

\bibliographystyle{splncs04}
\bibliography{main}

\end{document}